\documentclass{article}
\usepackage{spconf,amsmath,graphicx}
\usepackage{url}
\usepackage{multirow}
\usepackage{amssymb}
\newcommand{\tabincell}[2]{\begin{tabular}{@{}#1@{}}#2\end{tabular}}

\title{APB2Face: Audio-guided face reenactment \\with auxiliary pose and blink signals}
\name{Jiangning Zhang\(^1\), Liang Liu\(^1\), Zhucun Xue\(^2\), Yong Liu\(^1\)}
\address{\(^1\)Zhejiang University, Hangzhou, China \\
\(^2\)Wuhan University, Wuhan, China}

\usepackage{amssymb}
\begin{document}
%
\maketitle
\begin{abstract}    
Audio-guided face reenactment aims at generating photorealistic faces using audio information while maintaining the same facial movement as when speaking to a real person. However, existing methods can not generate vivid face images or only reenact low-resolution faces, which limits the application value. To solve those problems, we propose a novel deep neural network named \emph{APB2Face}, which consists of \emph{GeometryPredictor} and \emph{FaceReenactor} modules. \emph{GeometryPredictor} uses extra head pose and blink state signals as well as audio to predict the latent landmark geometry information, while \emph{FaceReenactor} inputs the face landmark image to reenact the photorealistic face. A new dataset $AnnVI$ collected from YouTube is presented to support the approach, and experimental results indicate the superiority of our method than state-of-the-arts, whether in authenticity or controllability.

\end{abstract}
\section{Introduction}
\label{sec:intro}
Audio and image are two commonly used signal transmission modes by humans, which both send overlapped conversation contents and rich emotional expression. When only hearing a familiar voice of a human, the subtle facial movements and the emotions of the people who is speaking could be imagined, especially for the mouth area. It is significant to study audio-to-face task that aims to generate the face image using audio as input, and this technique can be applied in many situations, such as virtual announcer, film-making
, game, etc.

Recently, many works focus on solving the audio-to-face task and have made some achievements. The methods mainly fall into two categories: parameter prediction and pixel-level generation.
The former usually predict parameters of the predefined model, which can control the face model in a post-processing way. The images generated in this way are of high quality, but it is time-consuming and does not work in real-time.
Pixel-level generation methods generally introduce the encoder to get a latent vector and then reenact face using the generative adversarial neural network (GAN)~\cite{goodfellow2014generative}, which require no post-processing and can run in an end-to-end manner.
Most recently, Tae-Hyun~\textit{et al.}~\cite{speech2face} propose the S2F model to predict the frontal face from the audio clip, and Amanda~\textit{et al.}~\cite{wav2pix} successfully reenact the special person by Wav2Pix model.
However, nearly all recent works can not generate high-quality faces only using audio as input and ignore other attributes such as head pose and eye blink, which results in stiff facial expressions. So in order to generate more photorealistic images, audio-independent control signals over head pose and eye blink are required.

\begin{figure*}[!ht]
    \centering
    \includegraphics[width=1.9\columnwidth]{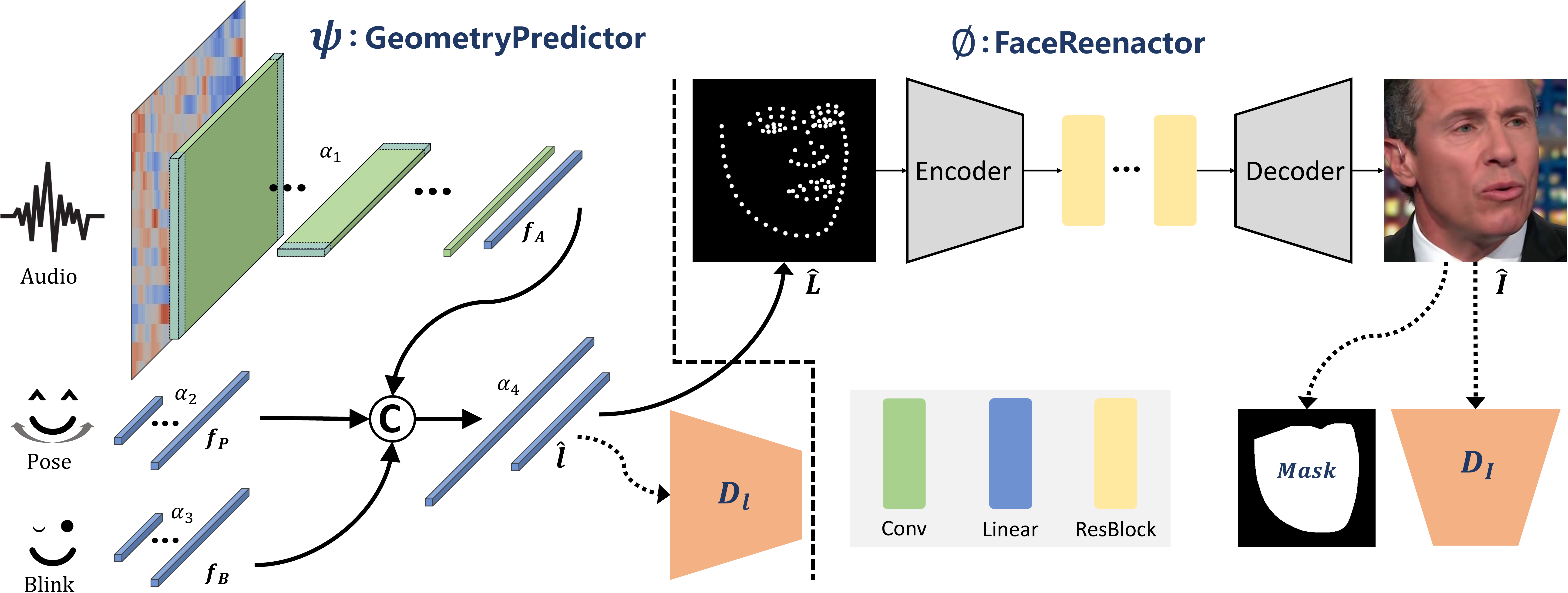}
    \caption{Overview of the proposed APE2Face network, which consists of a geometry predictor $\psi$ and a face reenactor $\phi$. The $\psi$ first encodes audio, head pose, and eye blink information as $\boldsymbol{f_A}$, $\boldsymbol{f_P}$, and $\boldsymbol{f_B}$, and then fuses them to regress the latent landmark geometry $\boldsymbol{\hat{l}}$. The discriminator $D_l$ is used to help boost performance and $\alpha_i(i=1,2,3,4)$ indicate network parameters. The $\phi$ leverages the transformed landmark image $\boldsymbol{\hat{L}}$ as input to reenact target face image $\boldsymbol{\hat{I}}$, whose characteristics should match the input audio, pose, and blink information. The discriminator $D_I$ helps to generate photorealistic images in the training phase.}
    \label{fig:architecture}
\end{figure*}

To solve aforementioned problems, we propose a novel \emph{APB2Face} structure as shown in Figure \ref{fig:architecture}. \emph{GeometryPredictor} module inputs audio, pose, and blink signals to regress latent \emph{landmark} that contains the facial geometry information (Thus we can use landmark image as input rather than a vector), and then \emph{FaceReenactor} reenacts the target image using the generated landmark. Such a two-stage design can not only generate photorealistic faces but also fuse pose and blink signals into audio. Our main contributions are as follows:

\romannumeral1) A novel \emph{APB2Face} structure is proposed to generate photorealistic faces from audio, pose, and blink signals.

\romannumeral2) A new dataset $AnnVI$ is created, which contains 6 announcers with synchronized audio, image, pose, landmark, and blink information.

\romannumeral3) Abundant experiments show that our approach can generate high-quality faces.

\section{Related Works}
\label{sec:rela}
\textbf{Generative Adversarial Networks}.
GAN method generally consists of two adversarial models: a generative model $G$ and a discriminative model $D$. The generator $G$ has to capture the data distribution and generate the realistic image, while the discriminator $D$ is used to discriminate generated fake images from $G$ and real images, in which way they can enhance each other's performance by playing a min-max game. Specifically, the generator $G$ with parameter $\theta_{g}$ maps input $\boldsymbol{z}$ from a prior distribution $p{_z}(z)$ to data space as $G(z; \theta_{g})$, and the discriminator $D$ with parameter $\theta_{d}$ outputs a probability value representing how much the input is from fake data space $G(z; \theta_{g})$ or real data space $\boldsymbol{x}$. The value function $V(G)$ and $V(D)$ are as follows:

\begin{equation}
\begin{aligned}
    \underset{G}\min~V(G) = \mathbb{E}_{\boldsymbol{z} \sim p_z(\boldsymbol{z})}[\log (D(G(\boldsymbol{z})) - 1).
\end{aligned}
\label{eq:G}
\vspace{-5pt}
\end{equation}

\begin{equation}
\begin{aligned}
    \underset{D}\min~V(D) &= \mathbb{E}_{\boldsymbol{z} \sim p_z(\boldsymbol{z})}[\log (D(G(\boldsymbol{z})))] \\
    &+ \mathbb{E}_{\boldsymbol{x} \sim p_{data}(\boldsymbol{x})}[\log (D(\boldsymbol{x}) - 1)],
\end{aligned}
\end{equation}

Since Ian~\textit{et al.} first proposed generative adversarial network~\cite{goodfellow2014generative} method in image generation task, many excellent jobs have come up~\cite{mirza2014conditional,isola2017image,zhu2017unpaired,karras2019style,huang2017arbitrary}. Considering the model size and the generation effect, we modified Pix2Pix model with a landmark input as our \emph{FaceReenactor}.

\textbf{Face Reenactment via Audio}. 
Some recent works reenact face by predicting parameters of the predefined face model\cite{karras2017audio,tian2019audio2face,cudeiro2019capture}.
Tian~\textit{et al.}~\cite{tian2019audio2face} directly use audio to regress blendshape parameters and then drive the predefined model.
Cudeiro~\textit{et al.}~\cite{cudeiro2019capture} design a VOCA model to drive the static 3D template. 
But they are not real-time for requiring post-processing, so another major approach is to directly generate pixel-level face images~\cite{suwajanakorn2017synthesizing, sadoughi2019speech, wav2pix, x2face}.
The state-of-the-art method~\cite{wav2pix} successfully reenacts faces of the special person by Wav2Pix model, which introduces a speech-conditioned GAN architecture. However, the quality of the generated images is not particularly good by this method, and it can not control generated attributes such as head pose and eye blink.

\vspace{-6pt}
\section{AnnVI Dataset}
\label{sec:dataset}
Consider that the current approaches usually use only audio information without other signals such as head pose and eye blink, we propose a new \emph{AnnVI} dataset with additional head pose and eye blink annotations which can complement latent head information that the audio cannot control. The dataset contains six announcers (three man and three women) and 23790 frames totally with corresponding audio clip, head pose, eye blink, and landmark information.

\textbf{Image processing:}
To crop out the sequence of faces from videos, we first extract video frames from original videos that are collected from YouTube at 1080P resolution, and then landmark geometry and head pose information of each frame are detected by~\cite{face}. Notice that we have manually checked the detection results of each image to ensure the reliability of the labeled information. After acquiring landmarks, we crop each face using its 1.4x minimum outer square and resize them to 256$\times$256 resolution.

\textbf{Audio processing:}
In this paper, all audios are firstly re-sampled to 44.1kHz and then we extract mel-frequency cepstral coefficients (MFCCs)~\cite{muda2010voice} features using the same method as~\cite{karras2017audio}.

\textbf{Pose and blink:}
Head pose information is detected by~\cite{face} that contains yaw, pitch, and roll angles. In AnnVI dataset, the value ranges of three components are $-0.354\sim0.196$, $-0.367\sim0.379$, and $-0.502\sim0.509$ respectively (radian).
As for eye blink, we define it as the height of the eye divided by the width, where the calculation scale is performed on the normalized 256$\times$256 resolution.

\vspace{-10pt}
\section{Method}
\label{sec:method}
Directly using a vector to reenact face generally limits the resolution and authenticity of the generated images. So as shown in Figure \ref{fig:architecture}, we propose a novel APB2Face structure, which consists of \emph{GeometryPredictor} and \emph{FaceReenactor}, to generate photorealistic pixel-level face from the audio input, along with additional head pose and eye blink control signals.
\textbf{Geometry predictor:}
The geometry predictor module contains three paths to extract various input information: audio, head pose, and eye blink.
For audio, we first use 5 convolutional layers to extract individual feature information and then use 5 convolutional layers to fuse features in the time dimension to get the feature $\boldsymbol{f_A}$.
For head pose and eye blink, 4 and 3 linear layers are used to extract features $\boldsymbol{f_P}$ and $\boldsymbol{f_B}$.
After that, we concatenate them and use 2 consecutive linear layers to predict finally landmark geometry. The process can be denoted as:
\begin{equation}
  \begin{aligned}
    \boldsymbol{\hat{l}} = & \psi_{\alpha_4}([\boldsymbol{f_A}, \boldsymbol{f_P}, \boldsymbol{f_B}]) \\
    = & \psi_{\alpha_4}([\psi_{\alpha_1}(Audio), \psi_{\alpha_2}(Pose), \psi_{\alpha_3}(Blink)])  \text{,}
  \end{aligned}
  \vspace{-2pt}
  \label{eq:generator}
\end{equation}
Note that the predictor is designed in a lightweight way, which can run in real-time: 260 FPS in CPU (with batch size 1, using i7-8700K @ 3.70GHz) and 823 FPS in GPU (with batch size 1, using 2080Ti).

During the training phase, an additional discriminator $D_l$ (containing 7 linear layers) is applied to provide stronger supervision beside L1 loss, which is designed to judge real or fake of generated landmarks. So the overall loss function $\mathcal{L}_{predictor}$ is defined as:
\begin{equation}
  \begin{aligned}
    \mathcal{L}_{predictor} = \lambda_{1}\mathcal{L}_{L_1} +
    \lambda_{2}\mathcal{L}_{D_l}  \text{,}
  \end{aligned}
  \label{eq:generator-loss}
\end{equation}
Where $\mathcal{L}_{L_1}$ and $\mathcal{L}_{D_l}$ indicate L1 loss and discriminator $D_l$ loss respectively, and $\lambda_{1}=100, \lambda_{2}=0.1$ in all experiments.
In this way, we can reduce the L1 loss of average generated landmarks from $1.500\pm0.064$ pixels to $0.666\pm0.034$, which greatly boosts the performance.

\textbf{Face reenactor:}
Given the predicted landmark $\boldsymbol{\hat{l}}$, we first plot it to the binary image according to its point coordinates, and then the resized landmark image $\boldsymbol{\hat{L}}$ (with 256*256 resolution) is fed into the \emph{FaceReenactor}. Each reenactor is associated with a special identity, which means that the input landmark image provides geometric information while the network itself contains the appearance information of the person. The process can be denoted as:
\begin{equation}
  \begin{aligned}
    \boldsymbol{\hat{I}} = & \phi(\boldsymbol{\hat{L}}) \text{,}
  \end{aligned}
  \label{eq:reenactor}
\end{equation}
Note that the reenactor can generate photorealistic high-resolution faces (256*256 resolution) and run in real-time: 75 FPS in GPU under the same setting with the predictor.
During the training stage, in addition to L1 and GAN losses~\cite{isola2017image}, we propose another mask loss function $\mathcal{L}_{L_{mask}}$. In detail, we first fill a black image with white in the smallest external polygon of the landmark and then expand the white area with the expansion algorithm to cover the face. Then L1 loss is applied to the white mask area, because we want to increase the weight of the face area that can help to generate the more photorealistic face. The overall loss function $\mathcal{L}_{reenactor}$ is defined as:
\begin{equation}
  \begin{aligned}
    \mathcal{L}_{reenactor} = \lambda_{1}\mathcal{L}_{L_1} +
    \lambda_{2}\mathcal{L}_{L_{mask}} + 
    \lambda_{3}\mathcal{L}_{D_I} \text{,}
  \end{aligned}
  \label{eq:generator-loss}
\end{equation}
Where $\mathcal{L}_{L_1}$, $\mathcal{L}_{L_{mask}}$, and $\mathcal{L}_{D_I}$ indicate L1 loss, mask loss, and discriminator $D_I$ loss~\cite{isola2017image} respectively, and $\lambda_{1}=100, \lambda_{2}=100, \lambda_{3}=1$ in all experiments.

\vspace{-10pt}
\section{Experiments}
\label{sec:exp}
In this section, we elaborate on the training details and show some experimental results to prove the effectiveness of the proposed method.

\textbf{Implementation details.}
For the geometry predictor, we use Adam~\cite{kingma2014adam} optimizer with $\beta_1=0.99$, $\beta_2=0.999$, and the learning rate is set to $3e^{-4}$. We train the module for 1000 epochs with batch size 32. For the discriminator $D_l$, we design 7 linear layers and the training setting is the same as the predictor.
For the face reenactor, we use Adam optimizer with $\beta_1=0.5$, $\beta_2=0.999$, and the learning rate is set to $2e^{-4}$. We train the module for 100 epochs with batch size 16 in a single GPU. For the discriminator $D_I$, we use PatchGAN proposed in~\cite{isola2017image} and the training setting is the same as the reenactor.

\begin{figure*}[!ht]
    \centering
    \includegraphics[width=1.8\columnwidth]{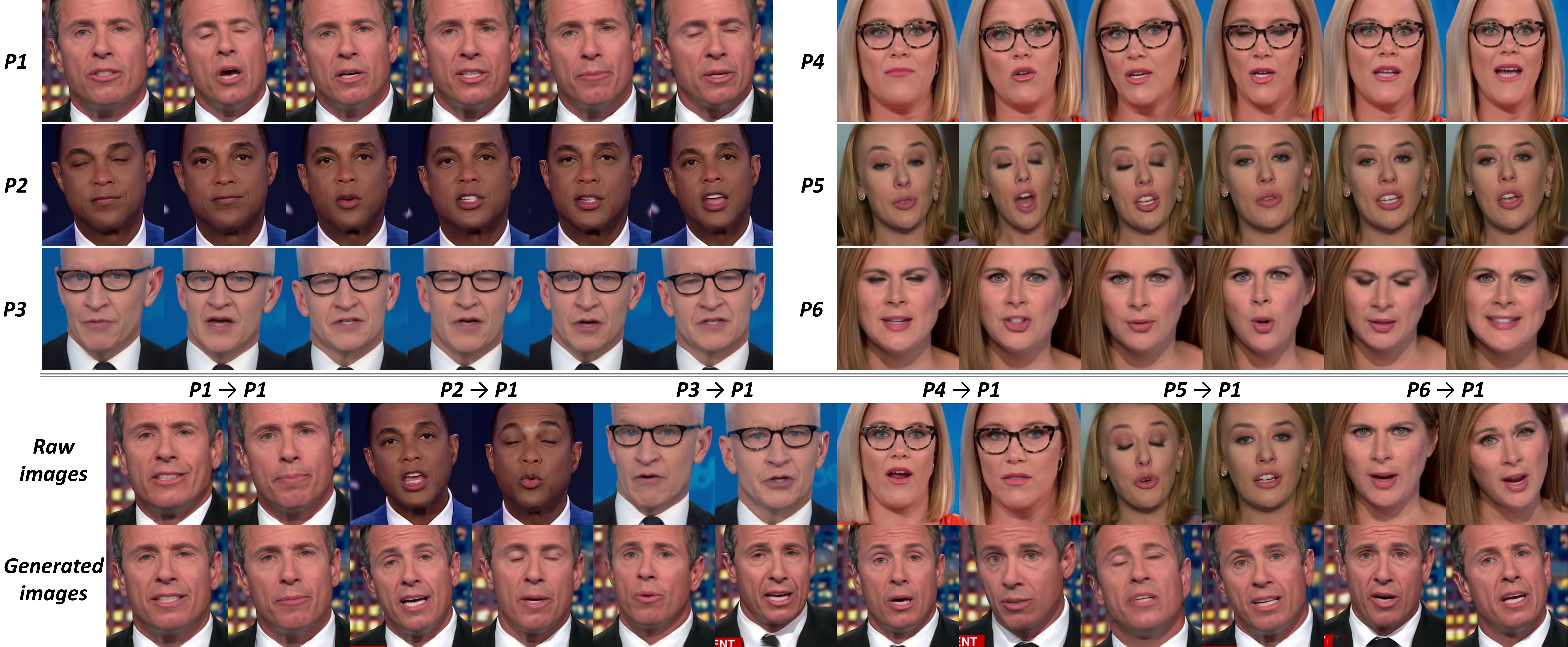}
    \caption{Face reenactment results of our proposed method. The top half is generated from individual audio input with pose and blink signals, while the bottom half is generated from the information of other persons. P1 to P6 represents different identities.}
    \label{fig:generation}
    \vspace{-20pt}
\end{figure*}

\textbf{Qualitative results.}
To visually demonstrate the effectiveness and superiority of our approach, we conduct some qualitative experiments as shown in Figure \ref{fig:generation}. The top half shows high-quality generation results of six persons, in which the input information of each identify is from himself/herself. To further evaluate the generalization capability of our model, input signals (audio, pose, and blink information) from $P1$-$P6$ are used to generate the identity $P1$ (Because of the limited space, we only show the results of $P1$), as shown in the bottom half in Figure \ref{fig:generation}. The generated faces match original faces in facial movements (especially mouth area), which indicates our approach can use audios from other persons to drive the given person.

\textbf{Quantitative results.}
As shown in the first two rows of the Table \ref{tab:metrics}, we choose generally used SSIM~\cite{wang2004image} and FID~\cite{Heusel2017GANsTB} metrics to quantitatively evaluate our approach, where results indicate that our model can generate photorealistic face images. The bottom four rows are evaluation results on our new defined \emph{Detection Rate} (DR), \emph{Average Landmark Error} (ALE), \emph{Average Pose Error} (APE), and \emph{Average Blink Error} (ABE) metrics, which are calculated on 1000 generated images for each identity. The results show that our method can reenact faces with an extremely high detection rate and low errors in landmark, pose, and blink information.

\begin{figure}[!ht]
\vspace{-15pt}
    \centering
    \includegraphics[width=1\columnwidth]{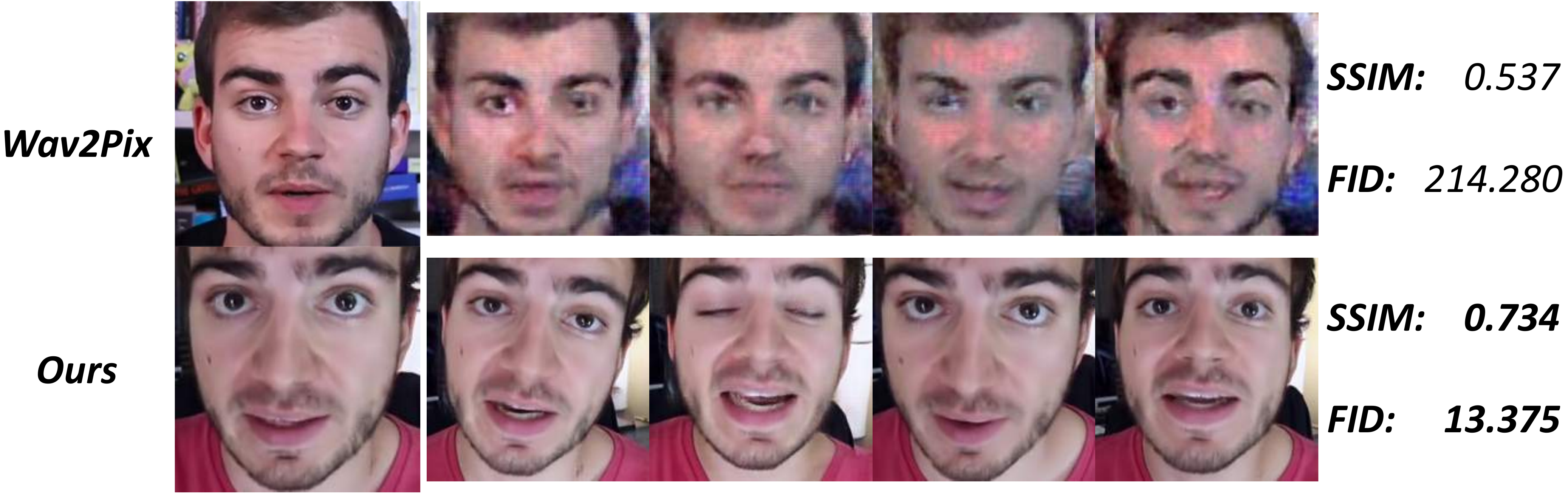}
    \caption{Comparison with the state-of-the-art method. The results of the first row are generated from Wav2Pix~\cite{wav2pix}, while the second row is ours.}
    \label{fig:comparision}
    \vspace{-20pt}
\end{figure}

\textbf{Comparison with the state-of-the-art.}
We further conduct a comparison experiment with the state-of-the-art \emph{Wav2Pix} in the same person (We use the video with the same identity as the original paper in this experiment, for the original author not providing the original data). As shown in Figure \ref{fig:comparision}, the result shows the superiority of our model no matter in vision (more photorealistic image generation) or metric evaluations (higher SSIM score and lower FID score).

\begin{table}[t] \small
    \begin{center}
    \caption{Quantitative evaluations of our proposed approach on six persons ($P1$-$P6$) in AnnVI dataset. The first two rows are SSIM and FID metric evaluation results, while the bottom four rows are evaluation results on our defined DR, ALE, APE, and ABE metrics. The upward arrow indicates that the larger the value, the better the performance, and vice versa.}
    \label{tab:metrics}
    \resizebox{240pt}{40pt}{
        \begin{tabular}{cccccccc}
        \noalign{\smallskip}
          \multirow{2}{*}{\textbf{Metric}} & \multirow{2}{*}{\tabincell{c}{\textbf{$P1$}\\(6400)}} & \multirow{2}{*}{\tabincell{c}{\textbf{$P2$}\\(6110)}} & \multirow{2}{*}{\tabincell{c}{\textbf{$P3$}\\(3450)}} & \multirow{2}{*}{\tabincell{c}{\textbf{$P4$}\\(3500)}} & \multirow{2}{*}{\tabincell{c}{\textbf{$P5$}\\(1230)}} & \multirow{2}{*}{\tabincell{c}{\textbf{$P6$}\\(3100)}} & \multirow{2}{*}{\tabincell{c}{\textbf{Avg}}} \\
          \\
        \hline
          SSIM $\uparrow$ & 0.764 & 0.843 & 0.879 & 0.786 & 0.761 & 0.758 & 0.799 \\
          FID  $\downarrow$& 4.600 & 6.514 & 5.743 & 14.960 & 17.408 & 21.944 & 11.862 \\
          \hline
          DR(\%)  $\uparrow$& 98.9 & 98.8 & 99.1 & 98.8 & 98.7 & 98.5 & 98.8 \\
          ALE  $\downarrow$& 1.237 & 1.119 & 1.303 & 1.587 & 1.464 & 1.861 & 1.429 \\
          APE  $\downarrow$& 0.0168 & 0.0143 & 0.0152 & 0.0250 & 0.0221 & 0.0233 & 0.0195 \\
          ABE  $\downarrow$& 0.0356 & 0.0370 & 0.0371 & 0.0502 & 0.0443 & 0.0437 & 0.0413 \\
        \hline
        \end{tabular}
        }
    \end{center}
\end{table}

\begin{figure}[!ht]
\vspace{10pt}
    \centering
    \includegraphics[width=1\columnwidth]{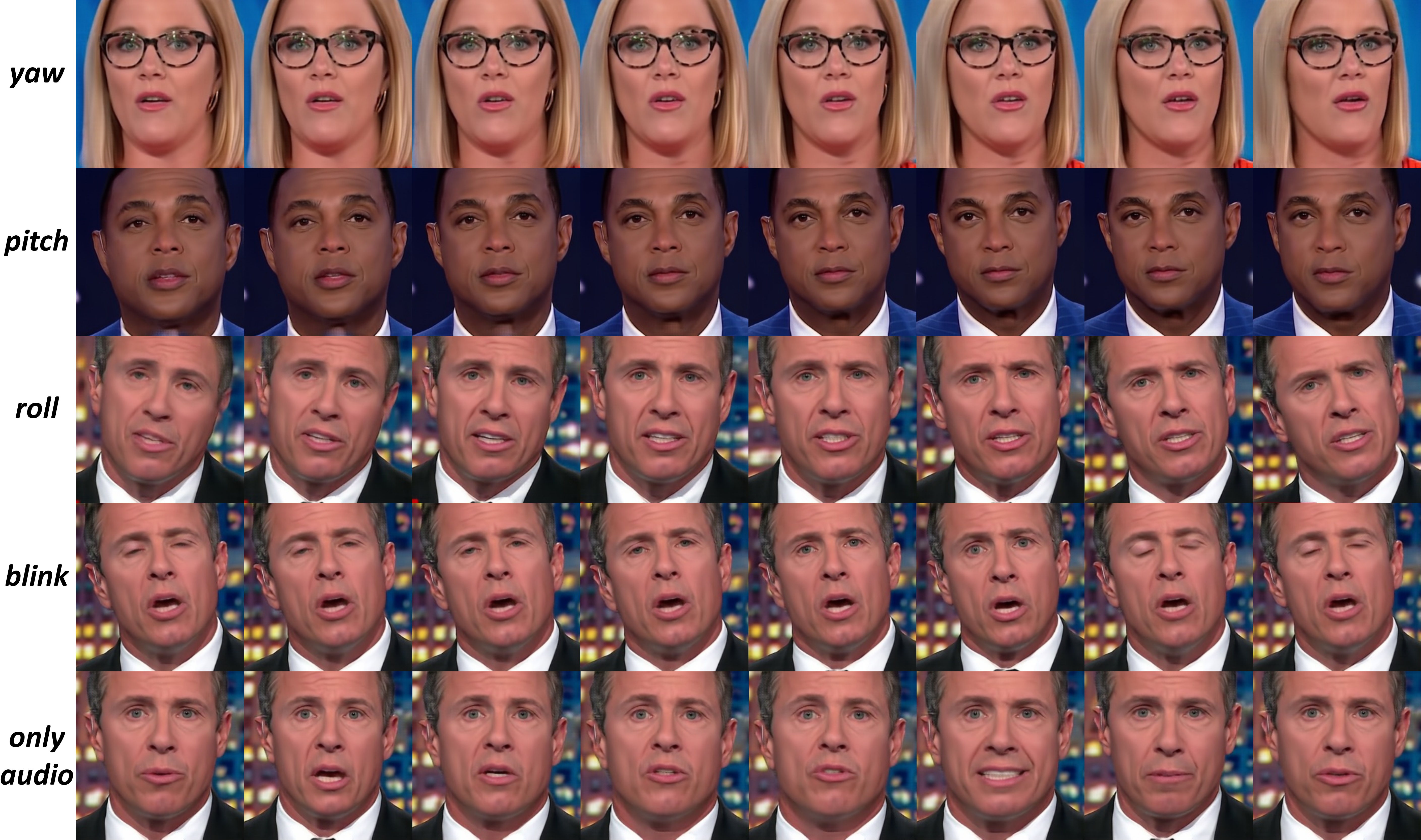}
    \caption{Decoupling experiments of the pose and the blink information. Images of the first four rows are under only one control variable, while the last row contains all variables.}
    \label{fig:decoupling}
\end{figure}

\textbf{Decoupling experiments.}
We also conduct a decoupling experiment to test the disentanglement of the audio, pose, and blink input information. As shown in Figure \ref{fig:decoupling}, the first four rows are generation results with only yaw, pitch, roll, and blink information changes respectively, and results indicate that our approach can well disentangle each component of the pose and blink signals, which gives us more room to control the generated image properties.
Results of the last row are generated with only audio signals (pose and blink are set to 0) like other methods~\cite{speech2face,wav2pix}, but the generated image quality of our method is visually better.

\section{Conclusions}
\label{sec:con}
A novel \emph{APB2Face} structure is proposed in this paper, which can use multi-information as control signals to reenact photorealistic faces. Specifically, we first design a \emph{GeometryPredictor} to regress the landmark geometry information from the audio, pose, and blink signals, and then we use \emph{FaceReenactor} to reenact target faces that match the input information. To verify the effectiveness of our approach and ensure the high quality of the results, we propose a new dataset named AnnVI which contains six announcers.

In the future, we will further extend our dataset as a stronger benchmark for the audio-to-face task.
In the meantime, a more powerful structure can be used to boost \emph{FaceReenactor} performance, so as to generate more photorealistic images.

\vspace{-10pt}
\section*{Acknowledgements}
This work is supported by the National Natural Science Foundation of China under Grant 61836015, 61771193.

\bibliographystyle{IEEEbib}
\bibliography{refs}

\end{document}